\documentclass[runningheads]{llncs}

\usepackage{eccv}

\usepackage{eccvabbrv}

\usepackage{graphicx}
\usepackage{booktabs}

\usepackage{colortbl}
\usepackage{array}
\usepackage{xcolor}
\usepackage{array}
\usepackage{graphicx}
\usepackage{float} 
\usepackage{tcolorbox}
\usepackage{algorithm, algorithmicx, algpseudocode}
\usepackage{multirow}
\usepackage[accsupp]{axessibility}  

\usepackage{hyperref}

\usepackage{orcidlink}

\begin{document}

\title{Decomposed Vector-Quantized Variational Autoencoder for Human Grasp Generation} 

\titlerunning{DVQ-VAE for Human Grasp Generation}

\author{Zhe Zhao\inst{1,2,3} \orcidlink{0009-0001-2722-5563}\and
Mengshi Qi\inst{1,2,3}\thanks{~Corresponding author: qms@bupt.edu.cn.} \orcidlink{0000-0002-6955-6635}\and
Huadong Ma\inst{1,2,3}\orcidlink{0000-0002-7199-5047}}

\authorrunning{Z.~Zhao et al.}

\institute{
State Key Laboratory of Networking and Switching Technology \and Beijing Key Laboratory of Intelligent Telecommunications Software and Multimedia \and Beijing University of Posts and Telecommunications, China
}

\maketitle

\begin{abstract}
  Generating realistic human grasps is a crucial yet challenging task for applications involving object manipulation in computer graphics and robotics. Existing methods often struggle with generating fine-grained realistic human grasps that ensure all fingers effectively interact with objects, as they focus on encoding hand with the whole representation and then estimating both hand posture and position in a single step. In this paper, we propose a novel Decomposed Vector-Quantized Variational Autoencoder~(DVQ-VAE) to address this limitation by decomposing hand into several distinct parts and encoding them separately. This part-aware decomposed architecture facilitates more precise management of the interaction between each component of hand and object, enhancing the overall reality of generated human grasps. Furthermore, we design a newly dual-stage decoding strategy, by first determining the type of grasping under skeletal physical constraints, and then identifying the location of the grasp, which can greatly improve the verisimilitude as well as adaptability of the model to unseen hand-object interaction. In experiments, our model achieved about \(14.1\%\) relative improvement in the quality index compared to the state-of-the-art methods in four widely-adopted benchmarks. Our source code is available at \url{https://github.com/florasion/D-VQVAE}.
\end{abstract}

\section{Introduction}
\label{sec:intro}

Generating how a human would grasp different given objects~\cite{jiang2021hand,karunratanakul2020grasping,karunratanakul2021skeleton,zheng2023coop} has already become crucial in many domains, including robotics, human-computer interaction, and augmented reality applications. Although 3D hand pose estimation~\cite{ge2018hand,zimmermann2017learning,boukhayma20193d,dibra2018monocular} and hand-object 3D reconstruction~\cite{hasson2019learning,fan2017point,tzionas20153d} have made significant strides in recent years, grasp generation extends beyond them and becomes imperative to have a finer-grained comprehension of the individual components of the highly articulated human hand. 

\begin{figure}[t]
    \centering
    \includegraphics[width=1\linewidth]{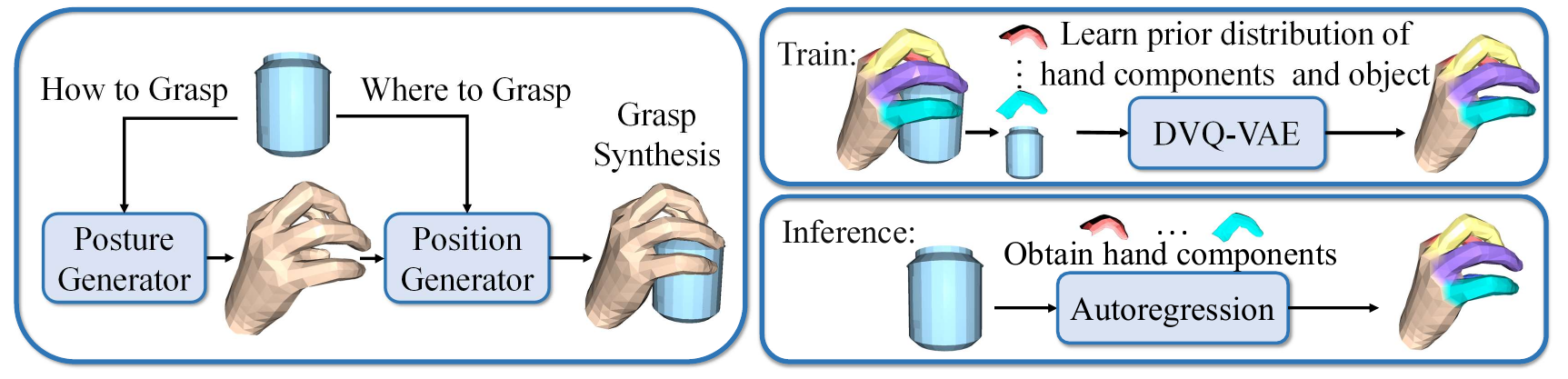}
    \caption{Illustration of the proposed grasp generation model. Initially, we utilize Decomposed VQ-VAE~(DVQ-VAE) to learn the prior distributions of the object and each hand component~(\ie, five fingers and the palm) during training. Specifically, we divide the decoding process into two stages hand posture and position generation. During inference, we perform autoregression using the object as a guide to obtain the realistically generated human grasp.}
    \label{fig:enter-label}
\end{figure}


To generate high-quality grasps, the main challenge lies in accurately modeling the interaction between the hand and objects. In recent years, numerous efforts~\cite{jiang2021hand,karunratanakul2020grasping,karunratanakul2021skeleton,zheng2023coop} have been adopted into this field, which is primarily focused on predicting the contact map to ensure the hand comes into full contact with the object. However, such methods often lead to generated unreasonable grasps that resemble touches rather than grasps~\cite{liu2023contactgen}. In addition, existing methods predominantly employ Conditional Variational Autoencoders (CVAE)~\cite{jiang2021hand,taheri2020grab,karunratanakul2021skeleton,karunratanakul2020grasping,liu2023contactgen,zheng2023coop}, while some have incorporated Generative Adversarial Networks (GANs)~\cite{corona2020ganhand,goodfellow2014generative}. But they utilize a continuous latent space to model all kinds of different grasps, which does not reflect the discrete and categorical nature of real-life human hands, resulting in poor diversity and fallacious grasps.
In contrast, Vector Quantized Variational Autoencoder (VQ-VAE)~\cite{van2017neural} can lead to a more controllable generation of grasps by directly utilizing discrete latent codes instead of randomly sampling from a Gaussian distribution. Therefore, we opt for VQ-VAE~\cite{van2017neural} as our core network and extend it in a new decomposed manner to take full advantage of each hand component and derive more diversity from the autoregressive inference. 

Another challenge is that existing methods cannot fit well into unseen or out-of-domain objects. Several research proposes test-time adaptation~\cite{jiang2021hand,karunratanakul2021skeleton}, which is an optimization-based refinement during inference by directly optimizing MANO parameters~\cite{romero2022embodied}. However, such approaches will lead to a significant time cost. To mitigate the time overhead during testing, we can break down the complex process of grasping into two distinct stages, \ie, predicting the hand pose by considering the possible grasp type and then finding the suitable position for the manipulated object.

In this paper, we present a novel Decomposed VQ-VAE~(DVQ-VAE) to generate human grasps that exhibit both diversity and authenticity when given an object. Our innovation lies in our model can partition the hand into multiple components for encoding them into several discrete latent spaces, and then executing grasp generation by the newly designed dual-stage decoding strategy. Specifically, we propose a part-aware decomposed architecture to decouple the hand into six distinct components including five fingers and the palm, each of which is encoded into its respective codebook. Furthermore, grasp generation occurs in two sequential steps by initially generating the posture of the hand followed by the determination of the grasp location.

Our main contributions can be summarized as follows:

\par\textbf{(1)} We propose a novel DVQ-VAE for human grasp generation, which is based on a part-aware decomposed architecture to encode multi-part of the hand and allows for progressively generating the complete hands. To the best of our knowledge, we are the first to propose such VQ-VAE~\cite{van2017neural} based framework for this specific task.

\par\textbf{(2)}  We introduce a new dual-stage decoding strategy to improve the quality of generated grasps for unseen objects, by gradually determining the grasp type under skeletal physical constraints and its position.

\par\textbf{(3)} We benchmark our proposed method on the four popular datasets, with  \(18\%\) relative improvements over the state-of-the-art approaches in the penetration volume. More importantly, we achieve nearly \(14.1\%\) relative improvement in the quality index metric.

\section{Related Work}
\label{sec:formatting}

{\bf Grasp Generation.}~With the development of virtual reality and robotics technology, significant efforts~\cite{corona2020ganhand,jiang2021hand,liu2023contactgen,karunratanakul2021skeleton,lv2024sgformer,lv2024disentangled,qi2019attentive,qi2018stagnet} have been devoted to generating the physically plausible and diverse grasps, most of which are based on the generative models, including GAN~\cite{goodfellow2014generative,corona2020ganhand,wang2022rgb,qi2020stc} and VAE~\cite{kingma2013auto,jiang2021hand,liu2023contactgen,karunratanakul2021skeleton,zheng2023coop}. A representative work is Jiang~\emph{et al.}~\cite{jiang2021hand} introduced a CVAE~\cite{sohn2015learning} to generate grasps, taking objects into account as conditions. This approach incorporated test-time adaptation by using ContactNet~\cite{jiang2021hand}. 
Furthermore, VQ-VAE~\cite{van2017neural} has demonstrated its superiority in multiple areas, such as image generation and 3D synthesis ~\cite{mittal2022autosdf,pi2023hierarchical}. 
However, it is noteworthy that, to the best of our knowledge, there exists no prior research that has applied VQ-VAE~\cite{van2017neural} to the field of grasp generation. Drawing inspiration from the work of Pi~\emph{et al.}~\cite{pi2023hierarchical}, who segmented the human body into five parts and encoded them into multiple codebooks for generating human motions, we extended this concept to encode the hand into multiple components. Through the utilization of VQ-VAE~\cite{van2017neural}, we encoded the data as discrete variables, thereby enabling us to acquire a prior distribution of hand-object interactions and then exploit it in the grasp prediction.

\noindent{\bf Hand-Object Interaction.}~Modeling hand-object interaction~\cite{romero2022embodied,hasson2019learning,brahmbhatt2020contactpose,grady2021contactopt,tzionas2016capturing,wang2020symbiotic,wang2021interactive,qi2021semantics} is a crucial task, but the considerable degrees of freedom in hand pose present significant challenges. A pioneer work is Romero~\emph{et al.}~\cite{romero2022embodied} proposed the parameterized hand model MANO to address the difficulties in reconstructing hand pose and shape. 
Different from modeling hand-object interaction assumes that both the hand and the object are visible in the input, we aim to predict a realistic grasp for the observed objects without seeing the hand.

 \section{Proposed Approach}
\label{sec:appro}

\begin{figure*}[t]
    \centering
    \includegraphics[width=1\linewidth]{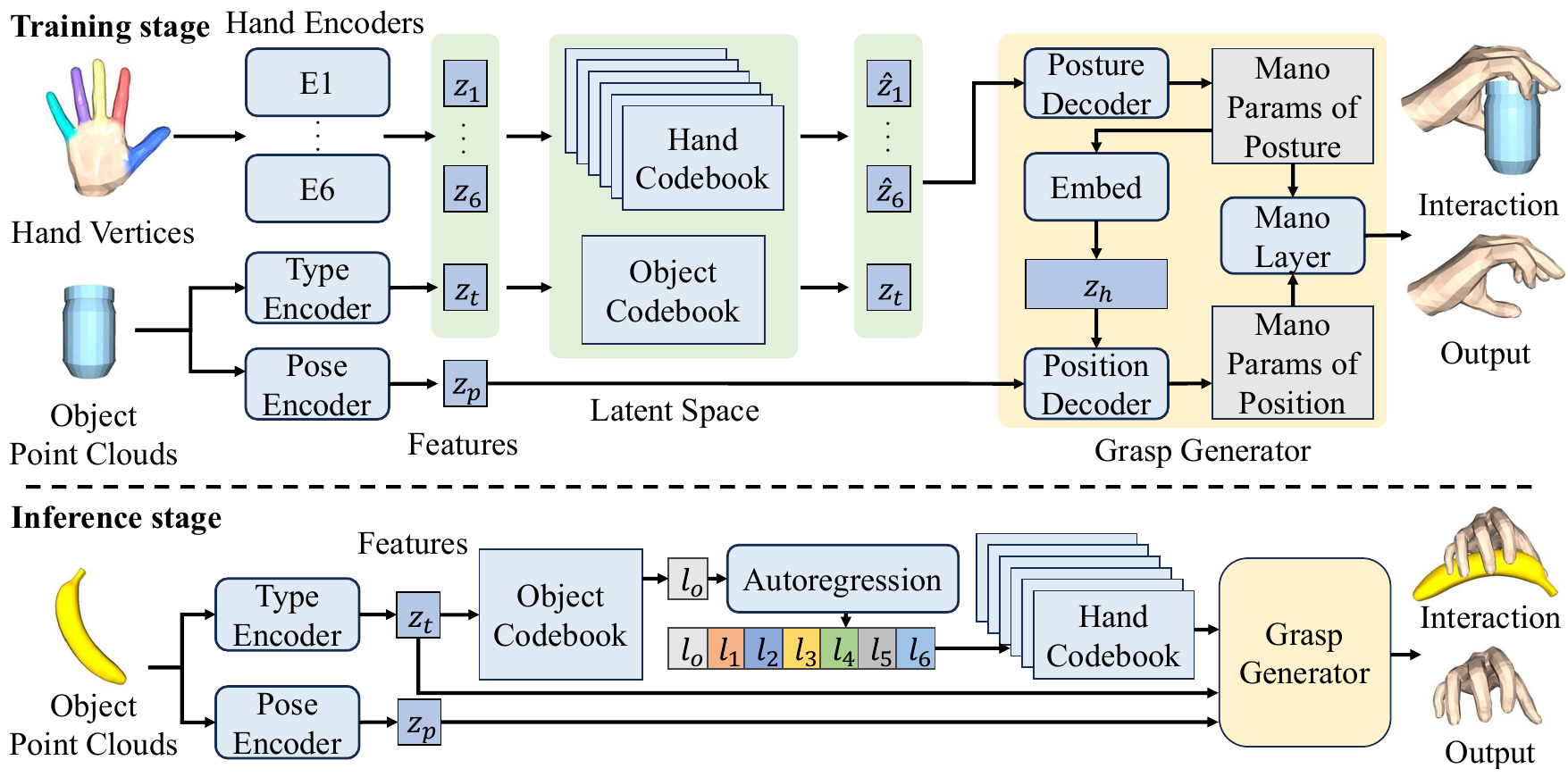}
    \caption{Overall architecture of the proposed DVQ-VAE model which is based on the encoder-decoder paradigm. During training, the model takes both hand vertices and object point clouds as inputs and maps them into discrete latent spaces consisting of seven codebooks~(\ie, one for object and six for hand) based on different hand components to generate hands. While, at the inference phase, we only use object point clouds as input to generate hands capable of grasping the given object. }
    \label{fig:network}
\end{figure*}

In our work, our goal is to generate a stable and physically plausible hand mesh for grasping by giving an object point cloud as input. To achieve this, we design a novel Decomposed VQ-VAE and divide the decoding process into two stages to obtain the final grasps, as illustrated in Fig.~\ref{fig:network}.


\subsection{Overview}
\label{subsec:arch}

{\bf Training.}~As shown in the top part of Fig.~\ref{fig:network}, given the sampled object points \(P^o\in \mathbb{R}^{N_o*3} \) and hand vertices \(P^h\in \mathbb{R}^{778*3} \) as input, we utilize two encoders to extract object features, denoted as \(z_t\) and \(z_p\). Simultaneously, we partition the hand vertices into \(N(N=6)\) segments, comprising five distinct finger components~(\ie, thumb, index finger, middle finger, ring finger, and little finger) and the palm, resulting in six distinct segments that are individually encoded to obtain hand features \(z_f=\{z_1, ...,z_N\}\). These hand features, along with \(z_t\), are projected into an discrete latent embedding space where \(N+1\) separate codebooks \(Z=\{Z_o,Z_1,...,Z_N\}\) are learned. During decoding, We use MANO~\cite{romero2022embodied} to generate hand mesh, and MANO~\cite{romero2022embodied} has several sets of parameters, \ie, \(M_{\alpha} \in \mathbb{R} ^{10} \) handles person-specific hand shape, \(M_{\beta} \in \mathbb{R} ^{45} \) manages joint rotations, \(M_{\gamma} \in \mathbb{R} ^{3}\) controls hand translation, and \(M_{\delta} \in \mathbb{R} ^{3} \) deals with hand rotation. Among these, \(M_{\alpha}\) and \(M_{\beta}\) are related to hand posture, and we combine them denoted as posture \(\hat{M}_{posture}\). While~\(M_{\gamma}\)~and \(M_{\delta}\) are related to hand position, and we combine them referred to position \(\hat{M}_{position}\). The returned closest matching vectors \(\hat{z}_f=\{\hat{z}_1, ...,\hat{z}_N\}\) of the hand features from \(Z\) are then concatenated with \(z_t\) and fed into the Posture Decoder to decode \(\hat{M}_{posture}\). Then, we concatenate the encoded result of \(\hat{M}_{posture}\), \ie, \(z_h\), with \(z_p\) and directed into the Position Decoder to obtain \(\hat{M}_{position}\). These two sets of MANO~\cite{romero2022embodied} parameters are then processed through the MANO~\cite{romero2022embodied} layer to derive the final hand mesh.

{\bf Inference.}~As shown in the bottom part of Fig.~\ref{fig:network}, our model utilizes the sampled object points \(P^o\in \mathbb{R}^{N_o*3} \) as input, and achieves object features,  \(z_t\) and \(z_p\) through the two object encoders. Then, \(z_t\) is matched with the nearest vector in the object codebook \(Z_o\) to obtain an index \(l_o\). Subsequently, we adopt an autoregressive model conditioned on \(l_o\) to generate a sequence of hand codebook indices \(l_h=\{l_1, ..., l_N\}\). These hand codebook indices are then matched in the learned hand codebooks to obtain the corresponding hand features \(\hat{z}_f=\{\hat{z}_1, ...,\hat{z}_N\}\). Finally, these hand features, along with \(z_t\) and \(z_p\), are fed into the grasp generator to produce the desired hand mesh. 

In our model, PointNet~\cite{qi2017pointnet} serves as the point cloud encoder for both objects and hands, and all decoders are MLPs. For the inference phase, we employ PixelCNN~\cite{van2016conditional} as the self-regressive model.


\subsection{Object Encoder}
\label{subsec:dis}


Different from existing methods~\cite{jiang2021hand,karunratanakul2020grasping,karunratanakul2021skeleton,liu2023contactgen,zheng2023coop,taheri2020grab}, we propose two object encoders: a type encoder and a pose encoder. The idea behind the type encoder is that all grasp types are uncountable and each of them is relevant to the shape of the given object, we can extract the $z_t$ to help learn an object codebook during the latent space embedding. And the codebook records these types and even clusters them into several common grasp categories so that our model can know how to grasp any input object. While, the pose encoder focuses on extracting features that aid in decoding grasp positions, which will determine where the hand should contact with the object. 


\subsection{Part-Aware Decomposed Architecture}
\label{sec:Part-Aware Decomposed Architecture}


Here we describe the proposed part-aware decomposed architecture of DVQ-VAE. Traditional VQ-VAEs~\cite{van2017neural} are typically used for image generation and employ a single codebook. This stems from the same codebook indices can be located in various positions of the image, while during autoregression the same codebook also can be used to infer the pixel at any position. However, in the case of human hands, the positions of different fingers are fixed. Therefore, we propose to extend VQ-VAE~\cite{van2017neural} to a part-aware decomposed architecture, \ie, encoding the object and the N parts of the hand as \(z_t\) and \(z_f\), respectively, and search for the vectors with the minimum Euclidean distance in their respective discrete codebooks during the latent embedding. To prevent the loss of object features, we use the discovered \(\hat{z}_f\) in combination with the original \(z_t\) as input for the grasp generator. In this scenario, the object's latent variable serves solely as a condition during autoregression. Thus, the training of the object codebook is conducted through unsupervised learning. In this context, to encourage \(z_t\) and \(z_f\) to closely approximate the matched vectors in the codebooks, the loss function \(\mathcal{L}_E\) for codebook training can be defined as follows:

\begin{equation}
\mathcal{L} _{h}= \sum_{i=1}^{N} \left \| sg(\hat{z}_i )- z_i \right \|^2_2+\beta  \sum_{i=1}^{N} \left \| sg( z_i)- \hat{z}_i\right \|^2_2,
\end{equation}
\begin{equation}
    \mathcal{L} _{o}=  \left \| sg(\hat{z}_t )- z_t \right \|^2_2+\beta   \left \| sg( z_t)- \hat{z}_t\right \|^2_2,
\end{equation}
\begin{equation}
    \mathcal{L}_E=\lambda_e\cdot(\mathcal{L}_h+\mathcal{L}_o),
\end{equation}
where
\begin{equation}
    \hat{z}_i = e_k,~\text{where}~k=argmin_j\left\|z_{i} - e_j\right\|_2,
    \label{eq1}
\end{equation}
\begin{equation}
    \hat{z}_t= e_m,~\text{where}~m=argmin_j\left\|z_t - e_j\right\|_2,
\end{equation}
where the operator \(sg(\cdot)\) signifies cessation of the gradient flow, impeding the propagation of gradients into its associated parameter, \(\beta\) is the hyperparameter set to 0.25, and \(\lambda_e\) is the hyper-parameter, \(e\) represents the embedding in the codebook. 


During inference, we follow VQ-VAE~\cite{van2017neural} to employ PixelCNN~\cite{van2016conditional} as our autoregressive model, and we use the object's codebook indices as both the condition and the initial sequence for PixelCNN~\cite{van2016conditional} to predict a sequence of codebook indices for the hand components.

\subsection{Dual-Stage Decoding Strategy}
\label{sec:Dual-Stage Decoding Strategy}


Previous methods~\cite{jiang2021hand,liu2023contactgen} improved grasp quality by generating 61 MANO~\cite{romero2022embodied} parameters in one step and optimizing these parameters over multiple steps, which resulted in significant time consumption. In order to overcome the weaknesses of single-step decoding, we design a dual-stage decoding strategy by generating the grasp posture and grasp position in sequence. In the spirit of decoupling, we divide the MANO parameters into two parts: position and posture, and we first generate the more numerous grasp posture parameters \(\hat{M}_{posture}\) and then the fewer position parameters \(\hat{M}_{position}\) by combining hand features with the object's characteristics.


1) Firstly, to generate a reasonable grasp during decoding, we utilize the L2 distance of the ground truth \(M_{posture}\) and the predictions:
\begin{equation}
\mathcal{L}_{posture} =\left \|M_{posture}-\hat{M}_{posture}  \right \|_2.
\end{equation}
Furthermore, we introduce physical constraints based on hand-skeletal dynamics. We extract skeletal key points \(J\) from the reconstructed hand and calculate angles between adjacent key points. From these angles, we generate gating information to correct grasps that do not conform to hand-skeletal dynamics:
\begin{equation}
\theta_i = \cos^{-1} \left( \frac{\overrightarrow{J_iJ_{i-1}} \cdot \overrightarrow{J_iJ_{i+1}}}{\|\overrightarrow{J_iJ_{i-1}}\| \|\overrightarrow{J_iJ_{i+1}}\|} \right),
\end{equation}
\begin{equation}
\hat{M}_{posture}=\hat{M}_{posture}+G(\theta) \odot T(\hat{M}_{posture}),
\end{equation}
where \(\theta\) represents the joint angles, \(\overrightarrow{J_iJ_{i+1}}\) represents the vector from key point \(J_i\) to \(J_{i+1}\), \(G(\cdot)\) denotes the network used to generate gating information, and \(T(\cdot)\) stands for the transformer layer used to produce correction values.

2) Secondly, for training the position decoder, we stop the gradient propagation at \(z_h\) and compute the L2 distance of \(M_{position}\) and hand vertices formulated as:
\begin{equation}
\hat{M}_{position}=Dec[sg(z_h),z_p],
\end{equation}
\begin{equation}
\mathcal{L}_{position} =\left \|M_{position}-\hat{M}_{position}  \right \|_2,
\end{equation}
\begin{equation}
\mathcal{L}_{v} =\left \|P^h-M(\hat{M}_{position},\hat{M}_{posture})  \right \|_2,
\end{equation}
where the operator \(Dec[\cdot]\) represents decoding through the position decoder, and the operator \(M(\cdot)\) represents passing through the MANO layer~\cite{romero2022embodied}. 
Then the reconstruction loss \(L_R\) is obtained by combining \(\mathcal{L}_{posture}\), \(\mathcal{L}_{position}\), and \(\mathcal{L}_{v}\):
\begin{equation}
\mathcal{L}_{R} =\lambda_h\cdot(\mathcal{L}_{posture}+\mathcal{L}_{position})+\lambda_v\cdot\mathcal{L}_{v},
\end{equation}
where $\lambda$ are hyper-parameters. To prevent the model from inadvertently learning hand position information before the position decoder, we center the hand vertices by subtracting their mean coordinates during training. 

\subsection{Optimization}
\label{sec:Loss Function}

Finally, our total objective loss consists of three parts, \ie, \(\mathcal{L}_E\) for the discrete latent embeddings in Sec.~\ref{sec:Part-Aware Decomposed Architecture}, \(\mathcal{L}_R\) for constraining the morphology of generated hand in Sec.~\ref{sec:Dual-Stage Decoding Strategy}, and \(\mathcal{L}_{contact}\) for constraining contact between the hand and the manipulated object that we will describe below. 

Following~\cite{jiang2021hand}, we use object-centric contact loss $\mathcal{L}_c$ and contact map consistency loss $\mathcal{L}_m$ to enhance the contact between the hand and the object formulated as:

\begin{equation}
    \mathcal{L}_c=\sum_{p_m\in P_m }^{} \min_{p_c\in P_c}\left | p_m-p_c \right |  ,
\end{equation}
\begin{equation}
    \mathcal{L}_m=\frac{\left |P_m  \right | \cap \left | \hat{P}_m  \right | }{\left |P_m  \right |},
\end{equation}

\noindent where the operator \(\left |\cdot \right |\) indicates the number of points in the point set. \(P_m\) and \(\hat{P}_m\) represent the point set of the ground truth and the predicted grasp contact map, respectively. \(P_c\) denotes the points on the hand that could potentially contact the object. $p_c$ and $p_m$ refer to each point in the point set \(P_c\) and \(P_m\), respectively.

We also employ penetration loss to enhance the physical reality of the generated grasp, formulated as: 
\begin{equation}
\mathcal{L}_p=  {\textstyle \sum_{p\in P_{in} }\left \| p-P^o_i \right \|^2_2  } ,
\end{equation}
where
\begin{equation}
P^o_i=\text{argmin}_{p_i\in{P^o} } \left \| p-p_i \right \|,
\end{equation}
where \(P_{in}\) is the subset of hand points that enter into the object. \(p\) and \(p_i\) denote as each point in the point set \(P_{in}\) and \(P^o\), respectively. Hence, we can obtain the contact loss as follows:

\begin{equation}
    \mathcal{L}_{contact}=\lambda_m\cdot\mathcal{L}_{m}+\lambda_c\cdot\mathcal{L}_{c} +\lambda_p\cdot\mathcal{L}_p.
\end{equation}
Where \(\lambda\) are hyper-parameters. 

Finally, we combine \(\mathcal{L}_{contact}\)  with \(\mathcal{L}_E\) and \(\mathcal{L}_R\) to form our overall loss function:
\begin{equation}
\mathcal{L}_{} =\mathcal{L}_{R}+\mathcal{L}_{E}+\mathcal{L}_{contact}.
\label{eq2}
\end{equation}
\section{Experiments}
\label{sec:exp}


\subsection{Datasets}
\label{subsec:dataset}


\noindent{\bf Obman Dataset~\cite{hasson2019learning}} constitutes a comprehensive compilation of manipulation interaction synthesized data, produced by GraspIt~\cite{miller2004graspit}. It encompasses a vast collection of 150,000 grasps across 2,772 distinct objects. Following~\cite{liu2023contactgen,jiang2021hand}, 141,550 of these grasps are designated for training purposes, while the remaining 6,285 are allocated for testing. 

\noindent{\bf HO-3D~\cite{hampali2020honnotate},~FPHA~\cite{garcia2018first} and GRAB~\cite{taheri2020grab} dataset.} Following~\cite{jiang2021hand}, due to the Obman dataset~\cite{hasson2019learning} containing the greatest variety of objects, we chose to train on the Obman dataset~\cite{hasson2019learning} but test on these three datasets. The test data can be regarded as out-of-domain objects used to evaluate the adaptability. Specifically, HO3D~\cite{hampali2020honnotate} contains 10 objects, FPHA~\cite{garcia2018first} contains 4 objects, and GRAB~\cite{taheri2020grab} contains 51 objects from 10 subjects. 

\subsection{Metrics}

For a fair comparison, following~\cite{liu2023contactgen,jiang2021hand,karunratanakul2020grasping,karunratanakul2021skeleton,hasson2019learning,tzionas2016capturing}, we evaluate the results in terms of the below metrics, of which the \emph{quality index} is our proposed new metric. 

\noindent{\bf 1) Contact Ratio~($\%$).} It calculates the ratio of grasps that can contact the given object in all generated grasps.

\noindent{\bf 2) Hand-Object Interpenetration Volume~($cm^3$).} It is the volume shared by the 3D voxelized models of the hand and object. Following~\cite{liu2023contactgen,karunratanakul2021skeleton}, We voxelize the mesh of both hand and object with size 0.1 $cm^3$. 

\noindent{\bf 3) Grasp Simulation Displacement~(Grasp Disp)~($cm$).} It quantifies the stability of the generated grasp when the simulated gravity is added, and we report the average displacement of the object's center of mass.

\noindent{\bf4) Entropy and Cluster Size.} Following~\cite{karunratanakul2021skeleton,liu2023contactgen}, we divide the generated grasps into 20 clusters using K-means and then measured the entropy and averaged cluster size to assess the diversity of generated grasps. Higher entropy values and larger cluster sizes indicate more diversity.

\noindent{\bf5) Time (\(s\)).} It evaluates the inference speed of the model in generating one batch of grasps. 



\noindent{\bf6) Quality Index.}~Note that we observe the visualized results with only low penetration or low physical simulation displacement does not necessarily indicate it is a high-quality grasp. For example, insufficient contact can reduce penetration but also decrease the stability of the grasp, while the severe penetrating of the hand into the object can reduce displacement in the physical simulator. Inspired by \cite{wang2024adaptive}, to more scientifically quantify the quality of grasping, we adopt a utility function:
\begin{equation}
Q =  a\cdot x+(1-a)\cdot y ,
\end{equation}
where \(a\) represents the weight, used to balance the gap between grasp displacement and penetration,\(x\) represents penetration
volume, \(y\) represents grasp disp. 
According to \cite{wang2024adaptive}, we measure the displacement and penetration for each grasp in the obman and grab datasets. And we calculate \(a=0.301\) using the same method as described in~\cite{wang2024adaptive}.
\subsection{Compared Methods}
\label{subsec:baseline}

We compare our proposed DVQ-VAE against the following state-of-the-art methods:~1)~{\bf GraspTTA~\cite{jiang2021hand}}: Trained on the Obman dataset~\cite{hasson2019learning} and utilizes CVAE~\cite{sohn2015learning} to generate grasps conditioned on the object, and further refines the generated grasps using ContactNet with test-time adaptation~(TTA). 2)~{\bf GraspCVAE~\cite{jiang2021hand}}: A variant of GraspTTA without TTA. 3)~{\bf ContactGen~\cite{liu2023contactgen}}: Trained on the Grab dataset~\cite{taheri2020grab} and utilizes CVAE~\cite{sohn2015learning} to generate contact maps of the object, and then optimizes the hand parameters based on them. 4)~{\bf Grasping Field~\cite{karunratanakul2020grasping}}: Trained on the Obman dataset~\cite{hasson2019learning} and utilizes VAE~\cite{kingma2013auto} to generate grasps based on the signed distance fields of the object and the hand.


\begin{table*}[t!]
    \centering
   \begin{tabular}{@{}c|c|cccc|cc|c@{}}
\toprule
Dataset & Method & \begin{tabular}[c]{@{}c@{}}Contact \\ Ratio\\~\((\%)\)↑\end{tabular} & \begin{tabular}[c]{@{}c@{}}Penetration\\  Volume\\ \((cm^3)\)↓\end{tabular} & \begin{tabular}[c]{@{}c@{}}Grasp \\ Disp \\ \((cm)\) ↓\end{tabular} &\begin{tabular}[c]{@{}c@{}} Time\\~\((s)\) ↓ \end{tabular}& \begin{tabular}[c]{@{}c@{}}Entr-\\opy~↑ \end{tabular}& \begin{tabular}[c]{@{}c@{}}Cluster \\ Size~↑\end{tabular} & \begin{tabular}[c]{@{}c@{}}Quality\\  Index↓\end{tabular} \\ \midrule
\multicolumn{1}{c|}{\multirow{5}{*}{\begin{tabular}[c]{@{}c@{}}HO-3D\\~\cite{hampali2020honnotate}\end{tabular}}} & 

\multicolumn{1}{c|}{GraspCVAE~\cite{jiang2021hand}} & \underline{99.60} & 7.23 & 2.78 & \multicolumn{1}{c|}{\textbf{0.0040}} & \textbf{2.96} & \multicolumn{1}{c|}{0.81} & \underline{4.12} \\

\multicolumn{1}{c|}{} & \multicolumn{1}{c|}{GraspTTA~\cite{jiang2021hand}} & \textbf{100} & 9.00 & \textbf{2.65} & \multicolumn{1}{c|}{19.67} & 2.87 & \multicolumn{1}{c|}{0.80} & 4.56 \\

\multicolumn{1}{c|}{} & \multicolumn{1}{c|}{ContactGen~\cite{liu2023contactgen}} & 90.10 & \underline{6.53} & 3.72 & \multicolumn{1}{c|}{119.4} & \underline{2.94} & \multicolumn{1}{c|}{\textbf{4.79}} & 4.57 \\
\multicolumn{1}{c|}{} & \multicolumn{1}{c|}{GraspingField~\cite{karunratanakul2020grasping}} & 89.60 & 20.05 & 4.14 & \multicolumn{1}{c|}{57.49} & 2.91 & \multicolumn{1}{c|}{3.31} &  8.93\\
\multicolumn{1}{c|}{} & \multicolumn{1}{c|}{\cellcolor[HTML]{E0E0E0}Our DVQ-VAE} &\cellcolor[HTML]{E0E0E0} 99.50 & \cellcolor[HTML]{E0E0E0}\textbf{5.36} & \cellcolor[HTML]{E0E0E0}\underline{2.75} & \multicolumn{1}{c|}{\cellcolor[HTML]{E0E0E0}\underline{0.14}} & \cellcolor[HTML]{E0E0E0}2.80 & \multicolumn{1}{c|}{\cellcolor[HTML]{E0E0E0}\underline{3.84}} & \cellcolor[HTML]{E0E0E0}\textbf{3.54} \\ \midrule
\multicolumn{1}{c|}{\multirow{5}{*}{\begin{tabular}[c]{@{}c@{}}FPHA \\ ~\cite{garcia2018first}\end{tabular}}} & \multicolumn{1}{c|}{GraspCVAE~\cite{jiang2021hand}} & \underline{98.98} & \underline{7.46} & \underline{2.97} & \multicolumn{1}{c|}{\textbf{0.0038}} & \underline{2.91} & \multicolumn{1}{c|}{0.81} & \underline{4.32 } \\
\multicolumn{1}{c|}{} & \multicolumn{1}{c|}{GraspTTA~\cite{jiang2021hand}} & \textbf{100} & 8.26 & \textbf{2.75} & \multicolumn{1}{c|}{19.29} & \textbf{2.93} & \multicolumn{1}{c|}{0.76} & 4.41 \\
\multicolumn{1}{c|}{} & \multicolumn{1}{c|}{ContactGen~\cite{liu2023contactgen}} & 94.00 & 10.43 & 3.64 & \multicolumn{1}{c|}{237.38} & 2.84 & \multicolumn{1}{c|}{\underline{2.88}} & 5.68 \\
\multicolumn{1}{c|}{} & \multicolumn{1}{c|}{GraspingField~\cite{karunratanakul2020grasping}} & 97.00 &29.78  & 5.47 & \multicolumn{1}{c|}{58.57} & 2.85 & \multicolumn{1}{c|}{2.36} & 12.79 \\
\multicolumn{1}{c|}{} & \multicolumn{1}{c|}{\cellcolor[HTML]{E0E0E0}Our DVQ-VAE} &\cellcolor[HTML]{E0E0E0} 97.96 & \cellcolor[HTML]{E0E0E0}\textbf{4.58} & \cellcolor[HTML]{E0E0E0}3.35 & \multicolumn{1}{c|}{\cellcolor[HTML]{E0E0E0}\underline{0.14}} &\cellcolor[HTML]{E0E0E0} 2.86 & \multicolumn{1}{c|}{\cellcolor[HTML]{E0E0E0}\textbf{3.53}} & \cellcolor[HTML]{E0E0E0}\textbf{3.72} \\ \midrule
\multicolumn{1}{c|}{\multirow{4}{*}{\begin{tabular}[c]{@{}c@{}}GRAB\\~\cite{taheri2020grab}\end{tabular}}} & \multicolumn{1}{c|}{GraspCVAE~\cite{jiang2021hand}} & 97.10 & \underline{3.54} & \underline{2.02} & \multicolumn{1}{c|}{\textbf{0.0041}} & \textbf{2.93} & \multicolumn{1}{c|}{0.81} &\underline{2.48} \\
\multicolumn{1}{c|}{} & \multicolumn{1}{c|}{GraspTTA~\cite{jiang2021hand}} & \textbf{100} & 5.05 & \textbf{1.74} & \multicolumn{1}{c|}{20.42} & 2.88 & \multicolumn{1}{c|}{0.93} & 2.74 \\
\multicolumn{1}{c|}{} & \multicolumn{1}{c|}{GraspingField~\cite{karunratanakul2020grasping}} & 74.80 &10.56  &3.80 & \multicolumn{1}{c|}{62.46} & \underline{2.91} & \multicolumn{1}{c|}{\underline{3.53}} &5.83  \\
\multicolumn{1}{c|}{} & \multicolumn{1}{c|}{\cellcolor[HTML]{E0E0E0}Our DVQ-VAE} & \cellcolor[HTML]{E0E0E0}\underline{98.60} & \cellcolor[HTML]{E0E0E0}\textbf{3.18} & \cellcolor[HTML]{E0E0E0}2.13 & \multicolumn{1}{c|}{\cellcolor[HTML]{E0E0E0}\underline{0.15}} & \cellcolor[HTML]{E0E0E0}2.83 & \multicolumn{1}{c|}{\cellcolor[HTML]{E0E0E0}\textbf{3.64}} & \cellcolor[HTML]{E0E0E0}\textbf{2.45} \\ \midrule
\multicolumn{1}{c|}{\multirow{4}{*}{\begin{tabular}[c]{@{}c@{}}Obman\\~\cite{hasson2019learning}\end{tabular}}} & \multicolumn{1}{c|}{GraspCVAE~\cite{jiang2021hand}} & 99.20 & \underline{4.32} & \textbf{1.81} & \multicolumn{1}{c|}{\textbf{0.0040}} & \underline{2.95} & \multicolumn{1}{c|}{1.50} & \textbf{2.57} \\
\multicolumn{1}{c|}{} & \multicolumn{1}{c|}{GraspTTA~\cite{jiang2021hand}} & \textbf{100} & 5.85 & \underline{2.06} & \multicolumn{1}{c|}{19.70} & \textbf{2.96} & \multicolumn{1}{c|}{1.50} & 3.20 \\
\multicolumn{1}{c|}{} & \multicolumn{1}{c|}{GraspingField~\cite{karunratanakul2020grasping}} & 74.62 &10.53  &3.81  & \multicolumn{1}{c|}{60.06} & 2.81 & \multicolumn{1}{c|}{\underline{2.33}}& 5.83 \\
\multicolumn{1}{c|}{} & \multicolumn{1}{c|}{\cellcolor[HTML]{E0E0E0}Our DVQ-VAE} & \cellcolor[HTML]{E0E0E0}\underline{99.82} & \cellcolor[HTML]{E0E0E0}\textbf{3.93} &\cellcolor[HTML]{E0E0E0} 2.70 & \multicolumn{1}{c|}{\cellcolor[HTML]{E0E0E0}\underline{0.14}} &\cellcolor[HTML]{E0E0E0} 2.90 & \multicolumn{1}{c|}{\cellcolor[HTML]{E0E0E0}\textbf{3.98}} & \cellcolor[HTML]{E0E0E0}\underline{3.07} \\ \bottomrule
\end{tabular}
    \caption{Performance comparison of our proposed DVQ-VAE and the state-of-the-art methods in terms of grasp generation on the 
 widely-adopted benchmarks, \ie, HO-3D~\cite{hampali2020honnotate}, FPHA~\cite{garcia2018first}, GRAB~\cite{taheri2020grab}, and Obman~\cite{hasson2019learning}. }
    \label{tab:table_1}
\end{table*}


\subsection{Implementation Details}
\label{subsec:imple}
We train our model on Obman dataset~\cite{hasson2019learning}, and sample \(N_o=3000\) points from the object mesh as input. We employ Adam optimizer with an initial learning rate of 1e-4 for 200 epochs, halving the learning rate at epochs 60, 120, 160, and 180, and set \(\lambda_e=10,\lambda_m=-50,\lambda_c=1500,\lambda_p=5,\lambda_h=0.1,\lambda_v=10\). For training PixelCNN~\cite{van2016conditional}, we use Adam optimizer with a learning rate of 3e-4 for 100 epochs. We implement our model based on Pytorch on a single NVIDIA RTX 3090 GPU, the training time is 1000 minutes.


\noindent
\begin{minipage}{.32\textwidth} 
  \centering
    
    \begin{tabular}{@{}l|l|@{}}
\toprule
Method & Score \\ \midrule
 Ours      &    \textbf{3.36}   \\
ContactGen~\cite{liu2023contactgen}&   3.25    \\
 GraspTTA~\cite{jiang2021hand}      &   3.23    \\ \bottomrule
\end{tabular}
    \captionof{table}{Average scores in human evaluation.}
    \label{tab:human}
\end{minipage}%
\hspace{2mm}
\begin{minipage}{.67\textwidth} 
\centering
\begin{tabular}{@{}l|c|c|c@{}}
\toprule
 &
  \begin{tabular}[c]{@{}l@{}}Contact \\Ratio~\((\%)\)↑\end{tabular} &
  \begin{tabular}[c]{@{}l@{}}Penetration\\ Volume~\((cm^3)\)↓\end{tabular} &
  \begin{tabular}[c]{@{}l@{}}Grasp \\Disp~\((cm)\) ↓\end{tabular} \\ \midrule
\(50\%\)Masked &
   99.60&
   5.66&
   2.80\\
\(90\%\)Masked &
   99.70&
   7.37&
   2.67\\ \bottomrule
\end{tabular}
\captionof{table}{Results after masking part of the object point cloud}
\label{tab:my-table}
\end{minipage}

\subsection{Results}
\label{subsec:result}

\begin{figure}[!t]
    \centering
    \includegraphics[width=1.0\linewidth]{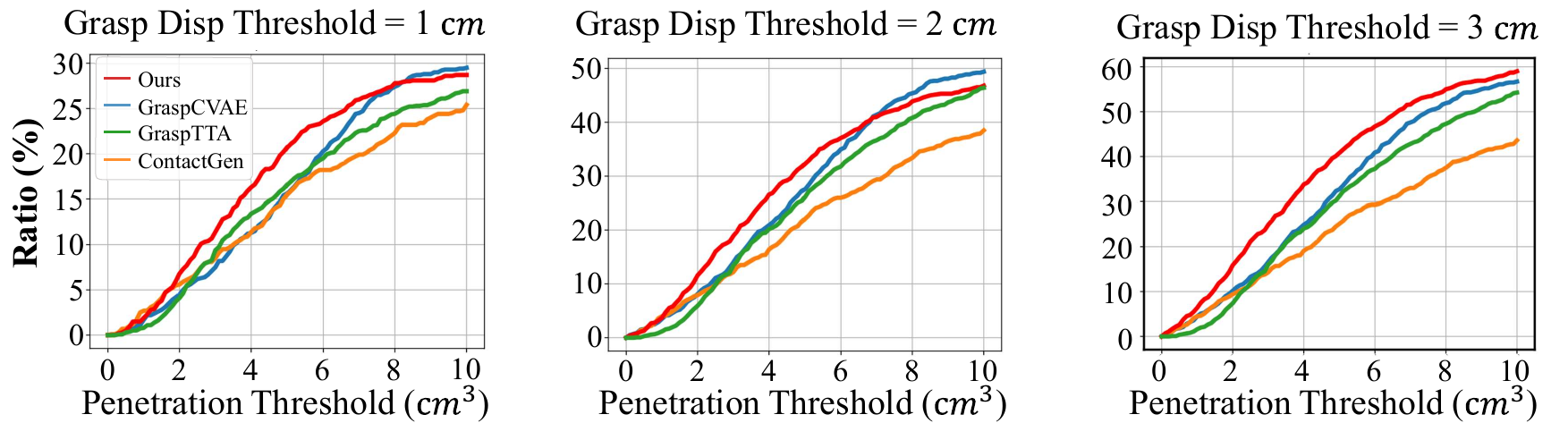 }
    \caption{Performance comparison of our method and other models in high-quality ratio w.r.t the penetration threshold for different models on the HO-3D dataset.}
    \label{fig:high-qualty}
\end{figure}

{\bf Quantitative Results.}~In experiments, we report the performance in Tab.~\ref{tab:table_1} of our proposed model and the other state-of-the-art methods which are all only trained on Obman~\cite{hasson2019learning} and test on HO3D~\cite{hampali2020honnotate}, Grab~\cite{taheri2020grab}, and FPHA~\cite{garcia2018first}. Note that the objects in HO3D, Grab, and FPHA are not present in the Obman training set and are never seen during the training process. As shown in the table, we can find our proposed DVQ-VAE can achieve the lowest penetration and grasp displacement approaches state-of-the-art benchmarks, demonstrating our proposed method allows for fine-grained controlling over touch interactions. Compared to the model~\cite{jiang2021hand} w.r.t quality index, our model shows a remarkable \(13.3\%\) relative improvement on HO-3D~\cite{hampali2020honnotate} dataset, \(13.7\%\) on FPHA~\cite{garcia2018first} dataset, and \(1.2\%\) on GRAB~\cite{taheri2020grab} dataset. And competitive results have also been obtained in terms of contact ratio and grasp disp used to assess grasping stability. Although GraspTTA~\cite{jiang2021hand} exhibits low grasp disp, its limited grasp diversity results in generating nearly identical grasp for a given object as mentioned in~\cite{liu2023contactgen}.  Moreover, our model achieved a relative improvement of \(22.6\%\) in the cluster size metric, suggesting the part-aware decomposed architecture we extended into VQ-VAE~\cite{van2017neural} allows our model to approach or even surpass the performance of leading methods in terms of grasp diversity. Meanwhile, we measured the ratio of high-quality grasps (both penetration and displacement are below certain thresholds) generated by each model, as shown in Fig.~\ref{fig:high-qualty} we report the ratio of each model on objects in the HO-3D dataset~\cite{hampali2020honnotate} as the penetration threshold varies from \(0~cm^3\) to \(10~cm^3\), and we can clearly see our model outperform the others most of times. These findings indicate the superiority and efficiency of our proposed DVQ-VAE.

{\bf Human Evaluation.}~We also conducted subjective experiments by inviting 10 human participants to evaluate the generated grasps of our DVQ-VAE, ContactGen~\cite{liu2023contactgen} and GraspTTA~\cite{jiang2021hand}. During the testing, given eight objects from the HO-3D~\cite{hampali2020honnotate} dataset, the models randomly generated 4 grasps for each object. Participants rated the grasps with a range from 1 to 5 in terms of \emph{penetration depth}, \emph{grasp stability}, and \emph{naturalness}, and the higher scores indicate the grasp is closer to a real human grasp. We report the results in Fig.~\ref{fig:human} and Tab.~\ref{tab:human}, it is evident that our proposed model produces grasps with the highest 5-point ratio and the lowest 1-point ratios, as well as achieving the highest average human evaluation score.

{\bf Robustness.}~We also tested the performance of our model when only partial point clouds of objects are provided, as shown in Tab.~\ref{tab:my-table}. Our model is capable of generating plausible results even with only partial point clouds, because of the powerful prior knowledge embedded in our codebook and autoregressive model. In contrast, ContactGen~\cite{liu2023contactgen} and GraspTTA~\cite{jiang2021hand} can generate grasps only from a fixed number of point clouds.

\noindent
\begin{minipage}{.47\textwidth} 
    \centering
    \includegraphics[width=1\linewidth]{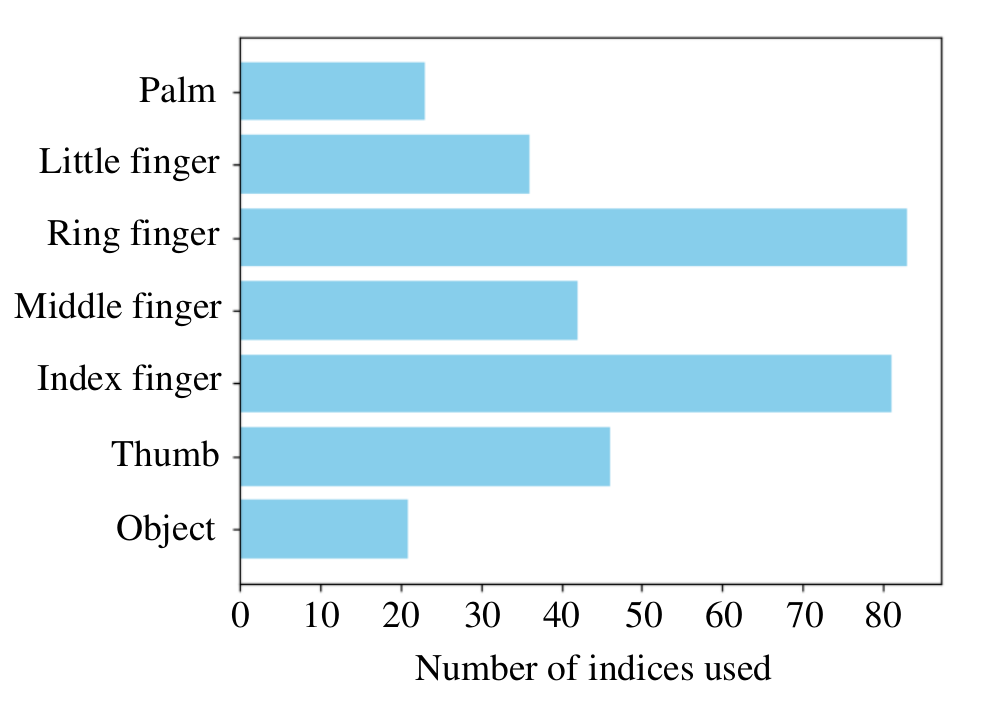}
    \captionof{figure}{The number of indices used in each codebook for our DVQ-VAE.}
    \label{fig:codebook}
\end{minipage}
\begin{minipage}{.52\textwidth} 
  \centering

    \includegraphics[width=0.9\linewidth]{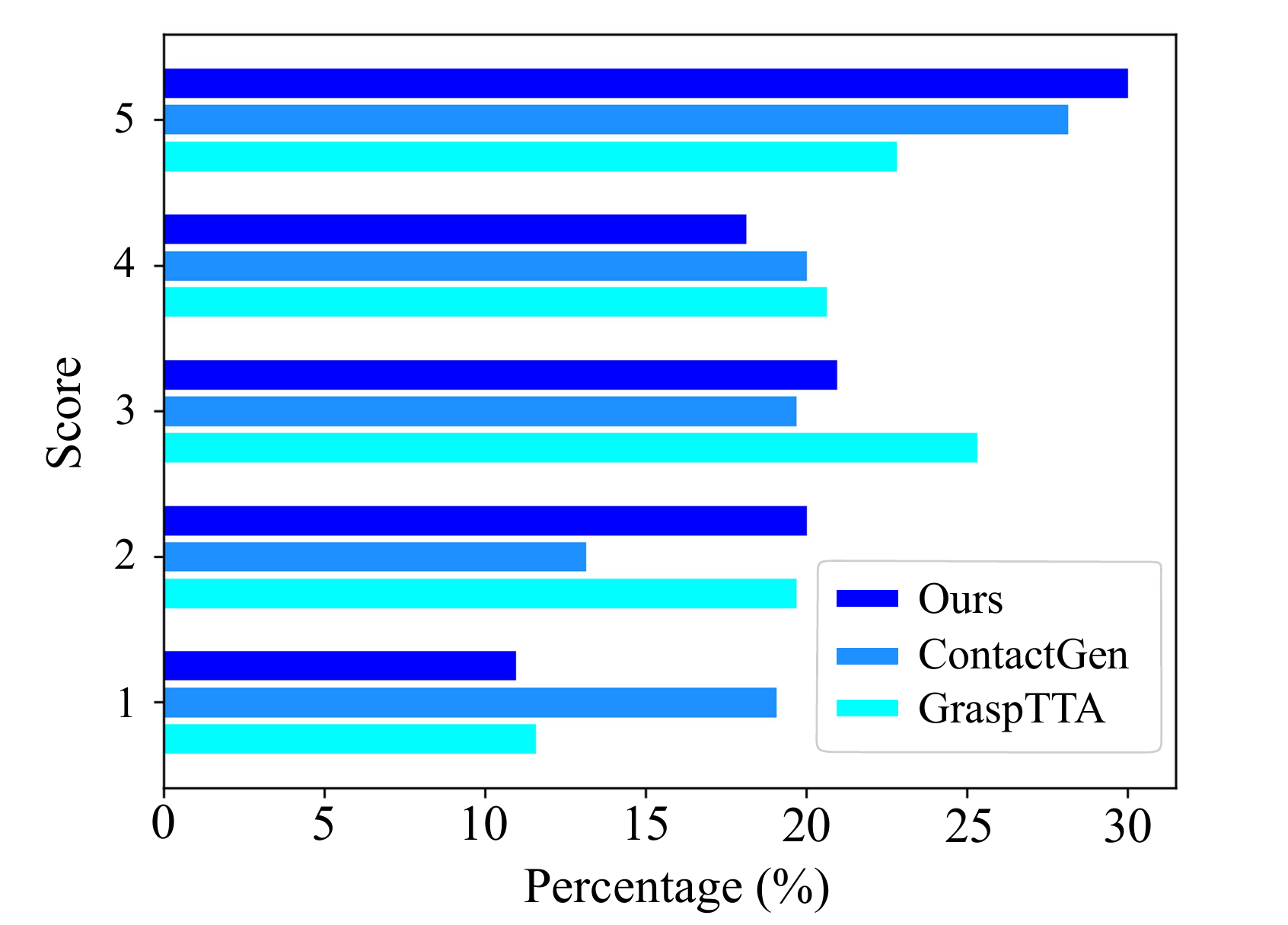}
    \captionof{figure}{The percentage of each score of each model in human evaluation.}
    \label{fig:human}

\end{minipage}

\begin{figure*}[!h]
    \centering
    \includegraphics[width=1.0\linewidth]{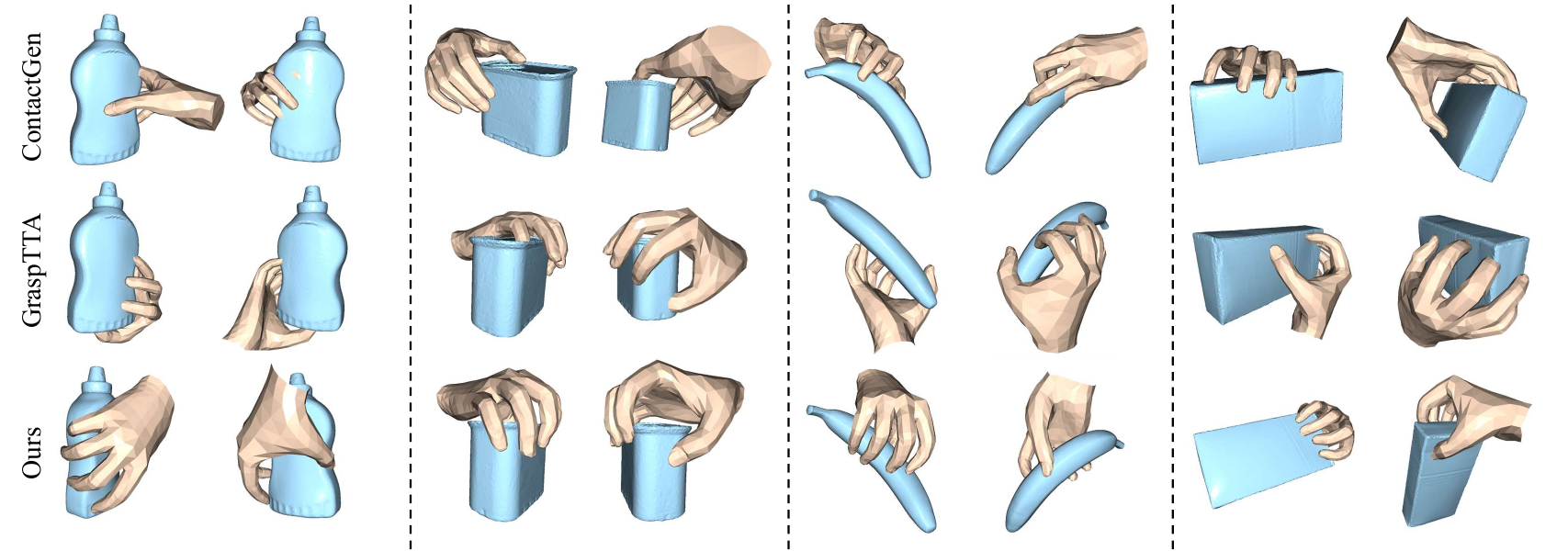}
    \caption{Qualitative results by comparing our DVQ-VAE with ContactGen~\cite{liu2023contactgen} and GraspTTA~\cite{jiang2021hand} on the HO-3D dataset~\cite{hampali2020honnotate}.}
    \label{fig:out-domain}
\end{figure*}

{\bf Inference Time.}~As shown in Tab.~\ref{tab:table_1}, we achieved a reduction of \(99.8\%\)/\(99.3\%\) time cost compared to ContactGen~\cite{liu2023contactgen} and GraspTTA~\cite{jiang2021hand}, respectively, demonstrates our proposed DVQ-VAE obtains faster grasp generation speed compared to methods using optimization methods during the generation period.

\begin{figure*}[!h]
    \centering
    \includegraphics[width=1.0\linewidth]{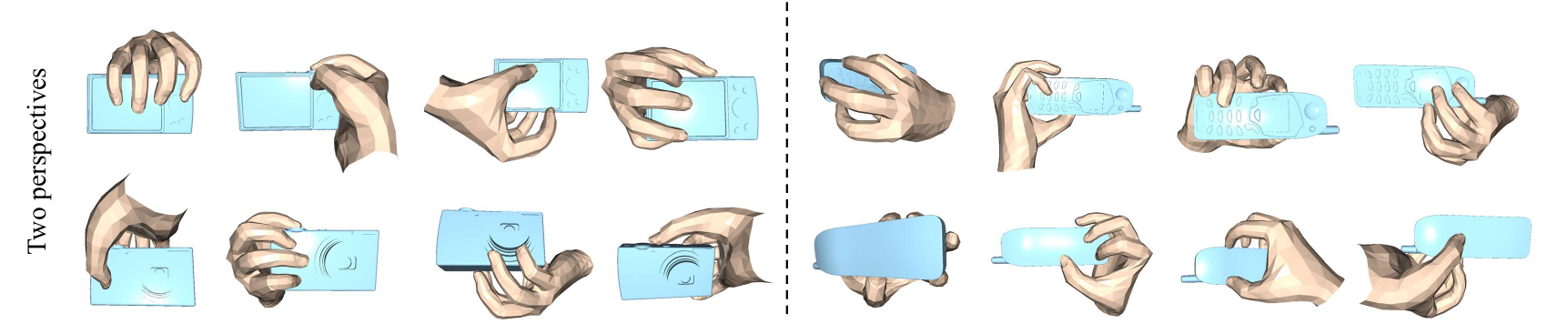}
    \caption{Diverse grasps generated by our DVQ-VAE for the same object in GRAB~\cite{taheri2020grab}.}
    \label{fig:diversity}
\end{figure*}

{\bf Qualitative Results.}~We visualize the grasps generated for different objects in various datasets, and each grasp is presented from two different perspectives. As Fig.~\ref{fig:out-domain} illustrated, we can clearly observe our model generates grasps with fewer penetrations and higher stability than others, showing strong adaptability when applied in such out-of-domain objects.
Furthermore, we also visualize the various types of grasps generated for the same object in Fig.~\ref{fig:diversity}, and it can be observed that our model can generate grasps with more diversity and different postures. Additionally, we show several visualized failure cases during testing in Fig.~\ref{fig:failure}, we can find our model may generate grasps with insufficient contact when applied to objects with complex geometric shapes. Enhanced representations of objects through the Signed Distance Field (SDF)~\cite{oleynikova2016signed} will be promising to mitigate this challenge. 

\noindent
\begin{minipage}{.48\textwidth}
    \centering
    \includegraphics[width=1\linewidth]{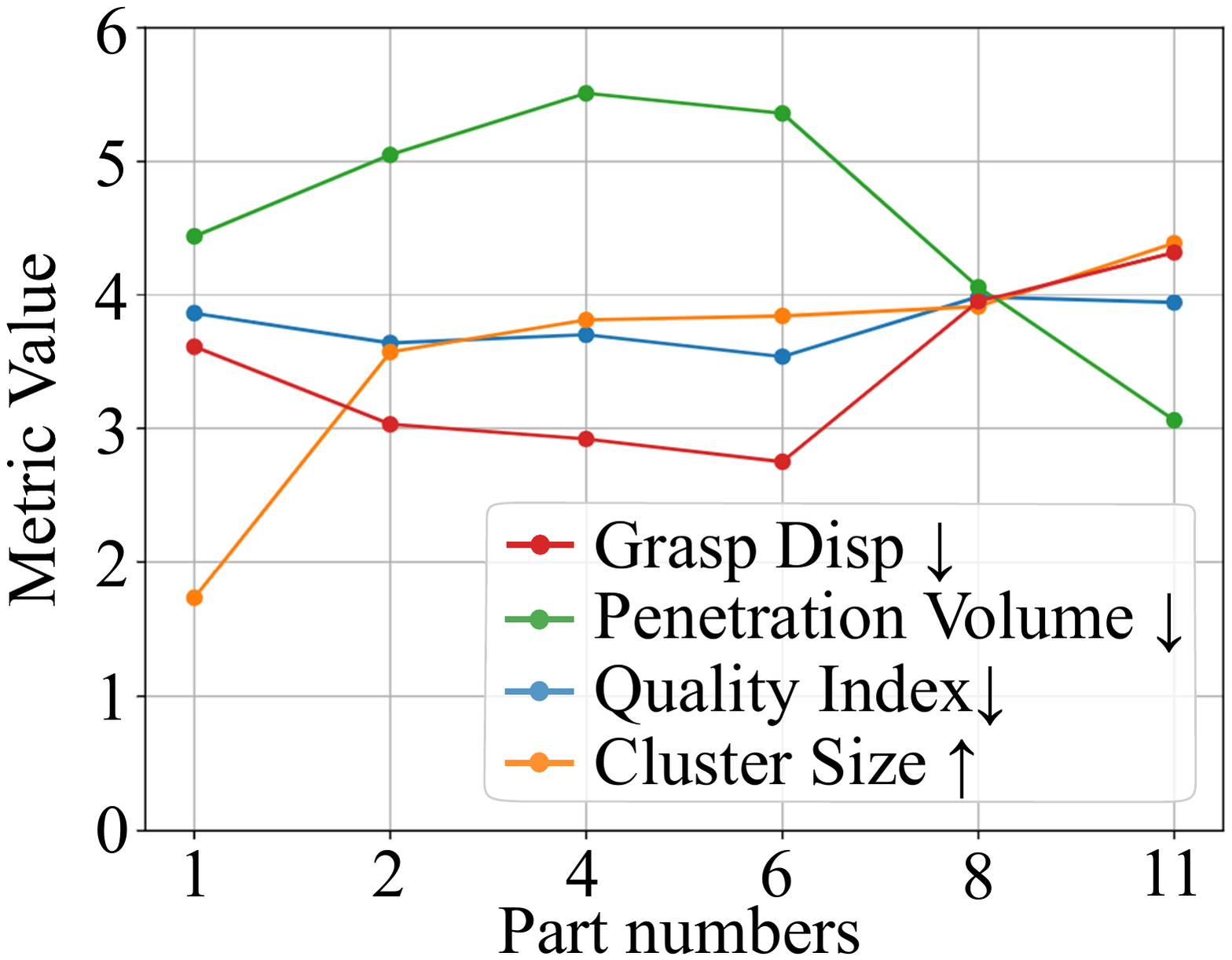}
    \captionof{figure}{Performance comparison of different parts used in our proposed DVQ-VAE on the HO-3D~\cite{hampali2020honnotate} dataset. We present the variation trend of evaluation metrics as the number of parts varied.}
    \label{fig:tendency}
\end{minipage}
\begin{minipage}{.52\textwidth}
    \centering
    \includegraphics[width=0.9\linewidth]{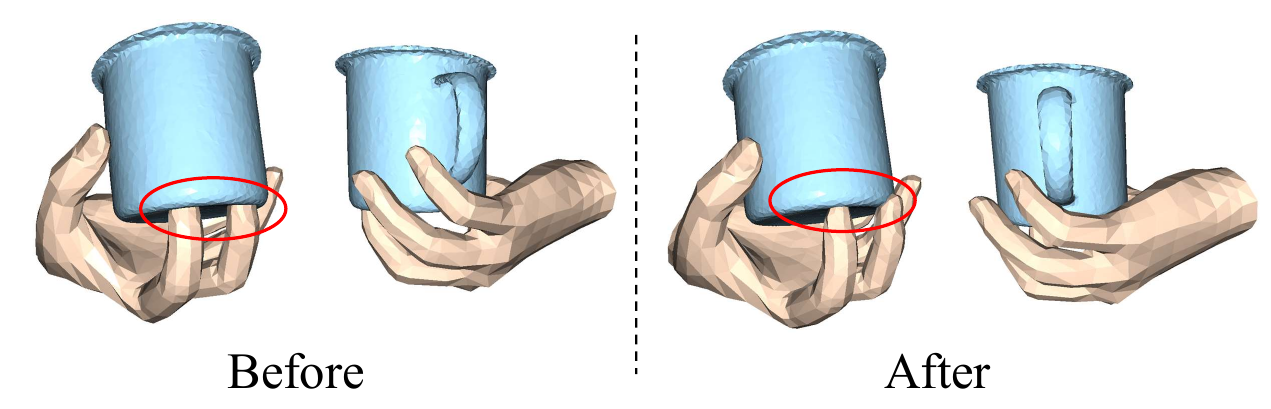}
    \captionof{figure}{Comparison of the generated grasp by the model with or without our proposed dual-stage decoding strategy, denoted as ``after'' and ``before'', respectively.}
    \label{fig:refine}
    \centering
    \includegraphics[width=1\linewidth]{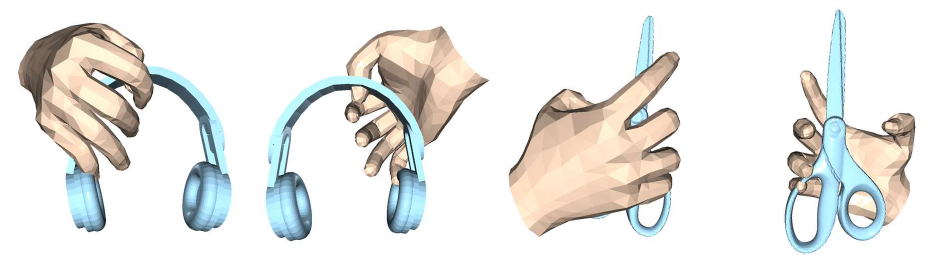}
    \captionof{figure}{Visualization examples of our model's failure cases. }
    \label{fig:failure}
\end{minipage}

\subsection{Ablation Study}
\label{subsec:ablat}

\begin{table*}[!t]
            \centering
            \begin{tabular}{@{}l|cc|cc|c@{}}
            \toprule
            Method & \begin{tabular}[c]{@{}l@{}}Penetration\\ Volume\\~\((cm^3)\)↓\end{tabular} & \begin{tabular}[c]{@{}l@{}}Grasp\\ Disp\\~\((cm)\) ↓\end{tabular} &\begin{tabular}[c]{@{}l@{}} Entro-\\py ↑ \end{tabular}& \begin{tabular}[c]{@{}l@{}}Cluster\\ Size↑\end{tabular} & \begin{tabular}[c]{@{}l@{}}Quality \\ Index~↓\end{tabular}  \\ \midrule
            VQ-VAE  & 6.67 & 7.21 & 2.82 & 1.69 & 7.05 \\
            VQ-VAE2  &\underline{5.18} & 9.87 & \textbf{2.96}& \underline{4.29} & 8.46 \\
            DVQ-VAE  & 10.88 & 4.98 & 2.94 & 4.24 & 6.76\\
            VQ-VAE+Dual-Stage(Two Encoders)  & \textbf{4.44} & 3.61 & 2.79 & 1.73& \underline{3.86}\\
            DVQ-VAE+Dual-Stage(One Encoder)  & 11.20 & 4.57 & \underline{2.95} & \textbf{4.31} & 6.57\\
            DVQ-VAE+Dual-Stage(Reverse) &7.56 &\underline{2.93} & 2.79&2.67 &4.32\\
            DVQ-VAE+Dual-Stage(Two Encoders)  & 5.36 & \textbf{2.75} & 2.80 & 3.84 & \textbf{3.54}  \\ \bottomrule
            \end{tabular}
    \caption{Performance comparison of DVQ-VAE and its variants in the ablation study. }
    \label{tab:table_ablation}

\end{table*}

\textbf{Object Encoder} We compare the results of encoding the given object with only one object encoder and with our proposed two object encoders in Tab.~\ref{tab:table_ablation}. We can find our proposed DVQ-VAE with two encoders outperforms the model with only one encoder across most of the metrics, which is attributed to the two encoders that can empower the object representation by decoupling the object's feature into type and pose parts. In addition, Fig.~\ref{fig:codebook} illustrates the distribution of index usage in various codebooks of our DVQ-VAE, it can be seen that objects are clustered into 21 categories based on the latent grasp type. From the usage of indexes, it can be observed that the degrees of freedom of the index finger and the ring finger are higher compared to other parts of the hand,  they have more hand posture prototypes.


\textbf{Part-aware Decomposed Architecture}
We ablate the decomposed architecture and show the performance in Tab.~\ref{tab:table_ablation}, and we can find the proposed architecture exhibits a relative improvement of \(22.4\%\) in the grasp stability compared to vanilla VQ-VAE~\cite{van2017neural}, along with a relative improvement of \(151\%\) in cluster size. This implies that our part-aware decomposed architecture not only enhances the diversity of generated grasps but also improves the quality of the generated grasps. Meanwhile, we compared the results in Fig.~\ref{fig:tendency} by utilizing DVQ-VAE with different numbers of parts on HO-3D~\cite{hampali2020honnotate}. It can be observed that dividing the hand into six components yields reasonable results by achieving a good trade-off considering all metrics. Furthermore, in the selection of the backbone network, we also tested VQ-VAE 2.0~\cite{razavi2019generating} structure that incorporates global features. However, as shown in Tab.~\ref{tab:table_ablation}, the results were not satisfactory. Instead, the autoregressive model PixelCNN~\cite{van2016conditional} used in our model can represent latent or implicit global hand features for reconstruction during inference.


\textbf{Dual-Stage Decoding Strategy.}~ In Tab.~\ref{tab:table_ablation}, we evaluate the effectiveness of our proposed dual-stage decoding strategy. Compared to models without the strategy, our full model can increase the contact ratio, and reduce penetration and displacement, thereby improving the overall grasping quality. Specifically, taking full use of this strategy results in \(45.2\%\) and  \(47.6\%\) relative increases in the quality index of our proposed DVQ-VAE and VQ-VAE~\cite{van2017neural}, respectively. We also tested decoding the position before the posture, but as indicated by the "Dual-Stage (Reverse)" results in the table, the outcome is not promising. Additionally, we also found that the position generation module in our trained DVQ-VAE can optimize the grasps generated by other models by improving the grasping positions. As shown in Fig.~\ref{fig:refine}, the optimized grasps exhibit reduced penetration, validating the effectiveness of our dual-stage decoding strategy.

\section{Conclusion}
\label{sec:conclu}

In this paper, we present Decomposed VQ-VAE for human grasp generation, including a part-aware decomposed architecture to account for different components of the hand with latent codebooks, and the dual-stage decoding strategy make the hand posture fit into the position in order. We have shown that each one of these components is important and helps outperform state-of-the-art methods. 

\section*{Acknowledgements} 
This work is partly supported by the Funds for the NSFC Project under Grant 62202063, Beijing Natural Science Foundation (L243027), the Innovation Research Group Project of the NSFC under Grant 61921003.


%
%
\bibliographystyle{splncs04}
\bibliography{main}
\end{document}